\documentclass[11pt,a4paper]{article}
\usepackage{lscape}
\usepackage{amsfonts, amsmath,amssymb}
\usepackage{graphicx}
\usepackage{graphics}
\graphicspath{ {./Bilder/} }
\usepackage{caption}
\usepackage[font=scriptsize]{caption}
\usepackage{epstopdf}
\usepackage{amsthm}
\usepackage{url}
\usepackage{ucs}
\usepackage[T1]{fontenc}\usepackage[utf8]{inputenc}
\usepackage{tabularx}
\usepackage{longtable}
\usepackage{ltablex}
\usepackage{hyperref}
\usepackage{color}
\usepackage{rotating}
\usepackage{lscape}
\usepackage{supertabular} 
\usepackage{natbib}
\begin{document}

\baselineskip=1.3\normalbaselineskip

\title{BERT based patent novelty search by training claims to their own description}

\author {Michael Freunek \thanks{University of Bern, Institute of Mathematics, Sidlerstrasse 5, 3012 Bern, Switzerland, Email: \texttt{michael.freunek@math.unibe.ch, m.freunek@gmx.de}} \and Andr\'e Bodmer \thanks{University of Berne, Institute of Economics, Schanzeneckstrasse 1, 3001 Bern, Switzerland, Email: \texttt{bodmerandre.ab@gmail.com, andre.bodmer@vwi.unibe.ch	}}}

\date{4. May 2021}
\maketitle

\begin{abstract}
\noindent
In this paper we present a method to concatenate patent claims to their own description. By applying this method, BERT trains suitable descriptions for claims. Such a trained BERT (claim-to-description-BERT) could be able to identify novelty relevant descriptions for pat-ents. 
In addition, we introduce a new scoring scheme, relevance scoring or novelty scoring, to process the output of BERT in a meaningful way. We tested the method on patent applications by training BERT on the first claims of patents and corresponding descriptions. BERT's output has been processed according to the relevance score and the results compared with the cited X documents in the search reports. The test showed that BERT has scored some of the cited X documents as highly relevant.

\bigskip

\noindent
\textsc{Keywords}: NLP, BERT, Supervised Learning, Transfer Learning, Downstream Task, Patent, Novelty Search, Prior Art, Search Report

\end{abstract}

\maketitle

\clearpage

\section{Introduction}

\noindent
Patent searches fulfill important tasks in the patent system. Patent applications are usually checked for novelty and inventive step in order to meet the requirements of the corresponding patent law. In the event of disputes between patent owners and alleged patent infringers, patents that have already been granted are usually checked again by the alleged patent infringer for legal validity. Relevant claims are searched again, as the examiner may have disregarded relevant prior art when the patent has been granted. Based on the classic Boolean methods, patent searches tend to be more and more extensive and complex. This is on one hand due to the rapidly growing number of new patents and other technical literature. On the other hand due to increasing technological interdisciplinarity. In addition, some claims include a large number of features and are difficult to search in total. With advancing development of artificial intelligence the number of projects and companies that investigate and offer search solutions, that are no longer based on pure Boolean search methods, is increasing. Different models and methods are used such as "bag of words", "words2vec", etc. combined with unsupervised and supervised methods.
With the publication of BERT and the availability via Huggingface, BERT enjoys great popularity in the field of natural language processing. BERT's applications in the field of patents are already known.

This work is inspired by the publication of \cite{Risch2020}. The authors provide a data set concerning patent examinations at the European Patent Office (EPO). Patent claims are matched against text passages, which, according to the EPO, are to be assessed with X (document of particular relevance) or A (document defining general state of the art). Own tests on this data set did not produce good results, even the idea of this data set is impressive. We suspect that the data set is not extensive enough for a given technology field and that the cited text passages may be too fragmented.

For this reason we have developed a new approach and trained BERT in a new way for novelty searches. The main idea is to relate or concatenate the patent claims to their own description. Assuming that the description of the claim were not taken from its own but from another patent, that other patent would be very likely novelty-destroying. If one wants to train BERT to identify a description of a claim that is destroying its novelty, then the description of a claim should provide the ideal training basis. Such a trained BERT should be able to identify novelty destroying descriptions for a claim in other patents or in other non-patent literature documents. In addition, there is an almost unlimited amount of data through this method, since virtually any patent can be used for training. It may be beneficial to train BERT specifically on a technology field or patent class, which can be easily implemented with the method. 

In this paper we describe this method in detail, apply BERT to five patent applications and compare the result with the cited X patents in the search reports\footnote{When we write about training BERT in the following, we mean fine tuning a pre-trained BERT.}. The method also includes a scoring, which we call relevance scoring or novelty scoring. To our knowledge, the method described in this work has not been published yet. The method presented here offers the following advantages:

\begin{itemize}
\item Data availability and data volume: Basically every patent  (in an appropriate language) is suitable for training. Almost any amount of data is available for any technology. This means that BERT can be trained very precisely for any searched claim.
\item Independence from published data sets. 
\item Data quality is very good and little dependent on subjective examiner evaluations (of course we assume that the examiners do excellent work). 
\item Handling of described procedure is quite simple and straightforward. 
\end{itemize}

\noindent
The paper is structured as follows: First, we identify related work in chapter \ref{related work}. Then we show our developed approach in chapter \ref{approach}: How we generate the data and which BERT model we use, how we practically apply the trained BERT for a novelty search, and how we evaluate the output of BERT in a relevance scoring process. In chapter \ref{experiment}, we apply the described method and train BERT on real patent data, apply the trained BERT to a pre-test, where BERT should identify the description corresponding to a claim in a group of patents. Finally, the trained BERT is applied to the actual test by searching novelty destroying prior art for five patent applications. The result is compared with the cited X patents in the search reports. We found that some of the cited X patents in search reports were given a high or even the highest priority by BERT. In chapter \ref{conclusion} we give some conclusions and possibilities for future research.\\

\section{Related Work}
\label{related work}

\noindent
There are several reasons for applying artificial intelligence (AI) techniques in the patent system, like increasing the efficiency in searching prior art for patent applications. \cite{Setchi2021} developed a platform to compare state of the art AI techniques in patent searches (feature extraction, query expansion, etc.) and concluded, that up to now, AI is not sufficient successful to fully automate the patent application and filing process. A similar finding is reported in \cite{Demey2020}. Further literature of applying AI to intellectual property as \cite{Aristodemou2018} identifies four categories: 1. knowledge management, 2. technology management, 3. economic value and extraction,  and 4. management of information.

Currently, BERT (Bidirectional Encoder Representations from Transformers) is one of the most popular models for tasks in natural language processing. BERT is based on the so called "self-attention mechanism" (cf. \cite{Vaswani2017}). "Self-attention" means, that different positions in a sentence or input sequence are related to each other. BERT is pre-trained on two tasks: the masked language model and next sentence prediction (cf. \cite{Devlin2018}). Applying BERT means fine-tuning the pre-trained BERT to a task like patent classification (cf. \cite{Sun2019} and \cite{Lee2020}). Patents are classified according to the IPC (International Patent Classification\footnote{\url{https://www.wipo.int/classifications/ipc/en}.}) and CPC (Cooperative Patent Classification\footnote{\url{https://www.epo.org}.}) by the patent offices according to the technical features, characterizing the invention.

\cite{Srebrovic2020} do research with pre-training BERT on patents. An advantage of training BERT to patents from scratch compared to a "standard" BERT is that the patent language typically differs compared to "standard" language. Many words, typically occurring in patents, are therefore not split during tokenization, which should increase the performance on patent NLP tasks. This is also true for tokens specialized to patents, like the claim token, whereby BERT can localize the text within a patent. BERT can be applied for patent prior art searches as well in combination with GPT2 (cf. \cite{Lee2020a}), or based on a data set matching claims of european patent applications to relevant or non-relevant text passages (cf. \cite{Risch2020}). In  a preliminary test, \cite{Risch2020} achieved an accuracy in the range 0.52 to 0.54 for two labels according to A or X patents. The application of transformer models on patent landscaping has been reported by \cite{Choi2019}. The authors use patent landscaping to identify related patents during research and development projects to avoid the risk of patent infringement.\\

\section{Approach}
\label{approach}

The method described below is based on the following idea: For nearly every claim existing in the patent data, there does exist a novelty matching text passage. Its the own description of the claims, that means the description of the invention in the patent itself. We match (concatenate) claims and its descriptions within the patents to train BERT, which learns to identify novelty relevant descriptions to claims. Such a trained BERT (claim-to-description-BERT) should have the capability to identify novelty relevant patents for a searched claim.\\

\subsection{Data Generation}
\label{DG}

BERT has a capacity of processing input lengths of size 512 tokens. Roughly estimated, a typical patent description amounts to several thousand words and therefore it is impossible to concatenate a claim to the whole description within one input sequence. By cutting the description into sections or pieces, we can concatenate a claim to such description piece. The question is, which description piece should be concatenated to the claim? Here, we make use of the following observation from the patent search practice: Most of the description passages fit very well to the claims within a patent. And we found in our research, that we can make a very good approximation to relate the whole patent to its claim by concatenating all sliced description pieces to the claim. Of course, this does not always apply in every single case, but from a statistical point of view this approximation should be very useful. 

In more detail, this works as follows: Each description-piece $k$ of patent $i$ is combined with the claim of the same patent $i$ according to the first label (here label 1): 

\begin{equation*}
\begin{split}
\textrm{Input}_{i,k} = \textrm{claim}({\textrm{patent}_i})\textrm{ <> } \textrm{SEP} \textrm{ <> } \textrm{description-piece}({\textrm{patent}_i})_k,
\end{split}
\end{equation*}
\\
where SEP means separation token. One should be aware to cut the description into pieces of a size, that the input length needs not to be truncated to fit the input length capacity of BERT.

Now, having the data for the first label, we need to complete the training data set by generating data with the second label (here label 0), where is no matching between claim and description piece. That means, we have to concatenate claims to description pieces, which are of no novelty relevance. Here, we make use of a second observation from the patent search practice: Most of the description pieces of a patent $j$ are of no novelty relevance of a searched claim of patent $i$, with $i \ne j$. Applying this observation, we can generate the second label data according to (here label 0):

\begin{equation*}
\begin{split}
\textrm{Input}_{i,j, k} = \textrm{claim}({\textrm{patent}_i})\textrm{ <> } \textrm{SEP} \textrm{ <> } \textrm{description-piece}({\textrm{patent}_j})_k,
\end{split}
\end{equation*}
\\
where $i \ne j$. Description pieces and claims are combined in a random manner, subject only to the limitation that claim and description piece belong to different patents.

By taking the same claims and descriptions pieces for the second label as for the first label, one can make sure, that both labels have the same frequency and BERT learns to distinguish between relevant and non-relevant description pieces for a claim. Of course, there is a probability of concatenating a claim randomly to a well fitting description piece of another patent, but at the end the effect a "confusing" during training should be very low. Finally, each input has the structure:

\begin{equation*}
\textrm{CLS} \textrm{ <> } \textrm{claim} \textrm{ <> } \textrm{SEP} \textrm{ <> } \textrm{text-piece} \textrm{ <> } \textrm{PAD}\textrm{ <> }\textrm{SEP} \textrm{ <> }\textrm{ <> } \textrm{label},
\end{equation*}
\\
where the length of the input is padded with PAD tokens during tokenization to the defined maximum input length, if the length of the concatenated claim and description piece is smaller than the maximum input length. Otherwise we get:

\begin{equation*}
\textrm{CLS} \textrm{ <> } \textrm{claim} \textrm{ <> } \textrm{SEP} \textrm{ <> } \textrm{text-piece} \textrm{ <> } \textrm{SEP} \textrm{ <> }\textrm{ <> } \textrm{label}.
\end{equation*}
\\
The maximum input length is not necessarily 512 token, but can also be set to a lower value beforehand. 
It is conceivable, that the size of the description pieces is "dynamically" adopted to the size if the concatenated claim, that the input length of the concatenated claim and description piece fits exactly the maximum input length. Beyond that, we expect the best result when the description pieces are at least as long as the claim or even longer. The claims include the technical features in a very compact form. A description is particularly relevant if it contains all the features of the claim, so it can be expected, that the size of the description piece has at least the same length as the claim.
\\

\subsection{BERT Model and Training}
\label{bert_model_training}

The following procedure is the usual procedure for training BERT. The first step is the tokenization of the training inputs. We applied the tokenizer {\bf bert-base-uncased, lower-case}. An important parameter here is the maximum input sequence length. BERT allows a maximum input length of 512 tokens. It is intuitively obvious to go close to this limit with this model, provided that the computer capacity allows it. In our experiment in chapter \ref{experiment} we set the input length to 500 tokens.

The maximum input length could be a significant limitation to BERT: Is the length of the searched claim in that range or even larger (which shouldn't happen often in practice), one would have the possibility of dividing the claim into subclaims. But the practice shows, that usually claims are much shorter than 500 words and there are good capabilities to concatenate a claim via separation token SEP to a description piece.

Hyperparameters like learning rate, batch size, number of epochs have to be adjusted and a BERT model has to be selected. The probably model of choice is the {\bf Bert For Next Sentence Prediction}. We experimented training the data with the BERT versions {\bf Bert For Next Sentence Prediction} and {\bf Bert For Sequence Classification} (both as {\bf bert-base-uncased}) and astonishingly got better results with {\bf Bert For Sequence Classification}. So in the following we will focus on that model. 

Before applying a trained model to the application of interest, there is a validation during training or testing after training to estimate overfitting or to estimate the performance of the model. We made the experience, that the significance of the validation of concatenated claim and description pieces, which are not trained explicitly, but in other combinations, is somewhat low. A better way for validating the training is using complete different patents. We did this after training BERT in a pre-test on 100 patents (see chapter \ref{pre-test}).\\

\subsection{Applying Trained BERT to Novelty Search}
\label{applying_trained_bert}

Once BERT is trained (fine-tuned), BERT can be applied to novelty searches for claims of interest. In our studies we focused training BERT on selected technology fields, normally represented by patent classes. The required computer capacity and the volume of data is therefore significantly lower. Also, it cannot be ruled out that a technology-specialized BERT is superior to a general trained BERT (trained to all or at least to several technology fields). The disadvantage, however, is that BERT must be trained specifically for a task, provided that the technology has not been trained before.

To prepare the input, we prepare the patents similar to the data generation procedure described in chapter \ref{DG}: The descriptions of the patents, which we want to analyze according to novelty relevance of the claim of interest, are sliced into description pieces. Again, the length of the description pieces has to fit to the length of the claim of interest and the chosen maximum input length to BERT. As before for training, we expect the best results here, when the description pieces are at least as long as the claim or even longer.

Then, the claim of interest - $\textrm{claim}_\textrm{oi}$ - is concatenated to the description pieces and we get the following structure (by neglecting the CLS and PAD token):
\begin{align*}
...\\
\textrm{Input}_{m}   &= \textrm{claim}_\textrm{oi}\textrm{ <> } \textrm{SEP} \textrm{ <> } \textrm{description-piece}({\textrm{patent}_j})_k \\
\textrm{Input}_{m+1} &= \textrm{claim}_\textrm{oi}\textrm{ <> } \textrm{SEP} \textrm{ <> } \textrm{description-piece}({\textrm{patent}_j})_{k+1} \\
\textrm{Input}_{m+2} &= \textrm{claim}_\textrm{oi}\textrm{ <> } \textrm{SEP} \textrm{ <> } \textrm{description-piece}({\textrm{patent}_j})_{k+2} \\
...\\
\textrm{Input}_{n}   &= \textrm{claim}_\textrm{oi}\textrm{ <> } \textrm{SEP} \textrm{ <> } \textrm{description-piece}({\textrm{patent}_{j+1}})_l \\
\textrm{Input}_{n+1} &= \textrm{claim}_\textrm{oi}\textrm{ <> } \textrm{SEP} \textrm{ <> } \textrm{description-piece}({\textrm{patent}_{j+1}})_{l+1} \\
\textrm{Input}_{n+2} &= \textrm{claim}_\textrm{oi}\textrm{ <> } \textrm{SEP} \textrm{ <> } \textrm{description-piece}({\textrm{patent}_{j+1}})_{l+2} \\
...
\end{align*}

\noindent
It is important to track the patent number for each description piece to finally link the search result of BERT with the corresponding patent.\\

\subsection{Postprocessing of BERT Results: Noise and Relevance Scoring}

Since a typical description in a patent amounts to several thousand words, a claim would be concatenated to tens of description pieces per patent. This, of course, can be affected by the length of the description pieces, what can be considered as hyperparameter. And maybe there is a variation of the length of the descriptions in patents dependent to the technology field / patent class or the country code of the patent. But finally there is a considerably number of concatenations per claim. The experiments have now shown, that BERT often labels a single description piece as relevant. In practice that means, it is very inefficient to regard every patent with at least one description piece labeled as relevant. In some sense, these single relevance labeling by BERT has to be understood as noise. But the great difference between novelty relevant and non-relevant patents is that the number of relevant labeled description pieces in relevant patents is much higher compared to non-relevant patents. For the the description of a claim within a patent, BERT  identifies nearly every description piece as relevant. To get rid of this noise and to distinguish relevant from at least less relevant patents, we calculated a relevance or novelty score in two ways.\\

\subsubsection{Relevance Scoring according to the Label}
\label{RS}

The relevance or novelty score $R(\textrm{claim}_\textrm{oi}, \textrm{patent}_i)$ for a claim of interest - $\textrm{claim}_\textrm{oi}$ - is calculated for a patent $i$ by simply the sum $\Sigma$ of relevant labeled claim and description piece concatenations of patent $i$ divided by the total number of the claim and description piece concatenations of patent $i$. The total number of claim and description piece concatenations of a patent $i$ is simply the sum of its label 0, $\Sigma\textrm{ }\textrm{label}\textrm{ }0(\textrm{patent}_i)$, and label 1 $\Sigma\textrm{ }\textrm{label}\textrm{ }1(\textrm{patent}_i)$. Because of the division, the scoring has the property of a density, as a result of which non-relevant patents with individual description pieces classified as relevant are given a low rating and relevant patents with several or even many description pieces classified as relevant are given a high rating. We finally get:

\begin{equation}
R(\textrm{claim}_\textrm{oi}, \textrm{patent}_i) = \frac{\Sigma\textrm{ }\textrm{label}\textrm{ }1(\textrm{patent}_i)}
{\Sigma\textrm{ }\textrm{label}\textrm{ }0(\textrm{patent}_i)+\Sigma\textrm{ }\textrm{label}\textrm{ }1(\textrm{patent}_i)}.
\end{equation}
\\
\noindent
Here we have arbitrarily assumed that label 1 classifies description pieces as relevant and label 0 as not relevant. In Chapter \ref{experiment} we will apply this scoring method instead of the method described in the following chapter \ref{SRS}, but there is no deeper reason for that.\\

\subsubsection{Relevance Scoring according to Sigmoid Values of Label 1}
\label{SRS}

An alternative way to calculate the relevance score $R(\textrm{claim}_\textrm{oi}, \textrm{patent}_i)$ for a claim of interest - $\textrm{claim}_\textrm{oi}$ - for a patent $i$ with the same density properties according to chapter \ref{RS} is simply to sum up all sigmoid values given for label 1. Even if BERT assigns label 0, i.e. the probability for label 1 is below 50 \%, the sigmoid value is added. The sigmoid values can be calculated with the logits output by BERT, yielding a relevance score for patent $i$ according to

\begin{equation}
R(\textrm{claim}_\textrm{oi}, \textrm{patent}_i) = \frac{\Sigma\textrm{ }\textrm{sigmoid}_{\textrm{label 1}}\textrm{ }(\textrm{patent}_i)}
{\Sigma\textrm{ }\textrm{label}\textrm{ }0(\textrm{patent}_i)+\Sigma\textrm{ }\textrm{label}\textrm{ }1(\textrm{patent}_i)},
\label{eq_RS2}
\end{equation}
\\
with the well known relation

\begin{equation}
\textrm{sigmoid}_{\textrm{label 0}}(\textrm{patent}_i) + \textrm{sigmoid}_{\textrm{label 1}}(\textrm{patent}_i) = 1.
\end{equation}
\\
It is conceivable that this method works better than the method according to the label in chapter \ref{RS}, since the assignment to label 0 and label 1 is not binary, but has probabilities. In this respect, according to the method in chapter \ref{RS}, no distinction is made between whether a label 1 assignment has only been made to 51 \% or 99 \%, but with this method, it is. In the numerator of equation \ref{eq_RS2}, the sigmoid values are also added up if the probability is below 50 \% and label 0 has been formally assigned. \\

\section{Experiment}
\label{experiment}

\noindent
We applied the method described above to a group of patent applications (reference patents) to answer the question, whether BERT is able to identify the X patents cited in the corresponding search reports. That means, does BERT identify the X patents as novelty relevant by assigning them a high relevance score according to chapter \ref{RS}, compared to non-relevant patents. Of course, the cited X patents do not necessarily have to be the best reference for a test by BERT, as there could be further or even better prior art. Also because we only train BERT on the first claims and we only search the first claims in the reference patents. If necessary, an examiner will select a patent according to the dependent claims, which we have not taken into account.
But this test is intended to give an initial indication of whether the method described above can be used reasonably and as a first benchmark returns the known, relevant patents. To this end, we performed the following procedure:

\begin{itemize}
\item[1:] Generation of 1059 training patents of label 0 and label 1 according to chapter \ref{DG} by a random selection of patents, which refer to the same patent class (here on IPC main-group level). These training patents form group 1.
\item[2:] Training BERT on the training patents (group 1).
\item[3:] Random selection of a second group (group 2) of 100 patents, as well chosen from the same patent class as group 1. This grop will be used for the pre-test according to chapter \ref{pre-test}.
\item[4:] Random selection of a third group of 1011 patents, as well chosen from the same patent class as group 1. This group has to contain the cited X patents. This group 3 we will call in the following to-be-searched group. BERT analyzes this to-be-searched group and score every patent according to the relevance or novelty score according to chapter \ref{RS}.
\item[5:] Identifying cited X patents in relevance and novelty score hierarchy.
\end{itemize}

Remark: We do intentionally not perform any text pre-processing.\\

\subsection{Data}
\label{the_data}

The novelty search of the first claims is performed for the following patent applications (reference patents): WO~2019/064177~A1, EP~3~593~317~A1, EP~3~382~640~A1, EP~3~438~918~A1 and EP~3~499~493~A1. For these five patent applications exists a search report, with the X citations given in table~\ref{t_reference_doc}.

\begin{table}[ht]
\centering
\begin{tabular}[t]{|l|l|}
\hline
\bf{Reference patent} & \bf{Cited X patent} \\
\hline
WO 2019/064177 A1 & US \, 2017/109857 A1 \\ 
& US \, 2011/194726 A1  \\
& WO 2005/029390 A1  \\
\hline
EP 3 593 317 A1 & WO 2016/025631 A1 \\
& US \, 2015/156369 A1  \\
\hline
EP 3 382 640 A1 & WO 2015/185944 A1  \\
& US \, 2005/193205 A1  \\
& US \, 2016/328398 A1 \\
\hline
EP 3 438 918 A1 & US \, 2014/029812 A1  \\
\hline
EP 3 499 493 A1 & US \, 2015/123887 A1  \\
\hline
\end{tabular}
\caption{Reference patents are given in the first column, the search report cited X patents in the second column. Novelty search is performend on the five reference patents.}
\label{t_reference_doc}
\end{table}%

\noindent
All reference patents are assigned to the IPC patent class G06T1/00. Most of the (but not all) cited X patents are classified in G06T1/00 and below.
For this reason, the 1'059 training patents (group 1), the 100 pre-test patents (group 2) and the 1011 to-be-searched patents (group 3) are as well randomly selected from G06T1/00 and below. The cited X patents were added to group 3 (if not already included randomly).  All randomly chosen patents in the groups 1-3 are US, EP, GB and AU patents in English language. It was checked, that BERT was not trained on the reference as well not trained on the cited X patents, although this is not a condition or limitation of the method presented. Between the randomly chosen training patents and the randomly chosen to-be-searched group (group 3) patents is an random overlap of 58 patents. There is no overlap to the pre-test (group 2) documents. The descriptions of the patents in the groups 1-3 were sliced into pieces of random size in the range of 100 to 200 words. Future research could investigate a dynamic slicing to fit exactly the size of the concatenated claim and description piece to the maximum sequence length, as described in chapter \ref{DG}. The maximum sequence length for BERT has been chosen to 500 tokens. Slicing the descriptions of the training patents (group 1) and the generation of label 0 and label 1 claim and description pieces `according to chapter \ref{DG}, yields 74'498 training input sequences. We split randomly 9'250 concatenated claim and description pieces for validation. As mentioned in chapter \ref{bert_model_training}: Although the concatenated claim description pieces are not trained, the significance of the validation set is rather low, since BERT is trained on the same claims and description pieces, although in other combinations. It would have been more reasonable to generate a validation set by taking claims and descriptions "not seen" by BERT during the training. This is done after the training in a pre-test, described in chapter \ref{pre-test}. Slicing the pre-test patents of group 2 and generation of the label 0 and label 1 by concatenating the claims to their description pieces (label 1) and the same claims and descriptions in a random manner to label 0 according to chapter \ref{DG}. This yields 3'582 input sequences of concatenated claim and description pieces. Slicing the descriptions for the novelty search of the to-be-searched patents of group 3 and generation the input sequences according to chapter \ref{applying_trained_bert} yields 40016 concatenated claim and description pieces. One note at the end: We transferred the descriptions of the patents via Excel. Excel has a limit of 32'767 symbols per cell. That does mean, that we did not train BERT in every case on the full patents. The cited X patent, not randomly included in the to-be-searched group, were completely adopted. If necessary, by splitting them into several cells. This does not necessarily mean, that all the other patents, truncated by Excel, get a lower or higher rating by BERT. The reason is the relevance scoring, which has the property of a density. That means adding text to be rated by BERT, positive and negative labeling should statistically be balanced to the rest of the text and therewith letting the scoring unchanged. Initial tests seem to confirm this assumption. However, this should be confirmed by further tests.\\

\subsection{BERT Training}

We trained BERT (Bert For Sequence Classification, Bert-base-uncased) for 2 epochs.\\

\subsection{Results}

\subsubsection{Pre-test on 100 Patents}
\label{pre-test}

In the pre-test we applied the trained BERT on 100 patents classified in IPC G06T1/00 (group 2). As described in chapter \ref{the_data}, in total we have 3'582 input sequences of concatenated claim and description pieces. In contrast to the novelty search as in the chapter \ref{test_bert_novelty_search}, the claims in these 100 documents are also concatenated with their own descriptions. This pre-test should therefore show, whether BERT is able to identify its own descriptions for the claims. The pre-test in the 100 patents yields a F1 score of 0.936. BERT very well identified the correct descriptions of the claims. In this pre-test we do not apply the relevance scoring according to chapter \ref{RS} or \ref{SRS}. The given F1 score refers to all 3'582 tested claim and description piece concatenations. If we were to use the scoring, the F1 score would be 100 \%. \\

\subsubsection{Test: BERT Novelty Search and X Patents} 
\label{test_bert_novelty_search}

The novelty search in the to-be-searched group (group 3) with 1011 patents of the trained BERT yields the results shown in table \ref{t_results}. The results are discussed in section \ref{discussion}.\\

\begin{table}[ht]
\centering
\begin{tabular}[t]{|l|l|c|}
\hline
\bf{Reference patent} & \bf{Cited X patent} & \bf{Pos. relevance score} \\
\hline
WO 2019/064177 A1 & US \, 2017/109857 A1  & 198\\ 
& US \, 2011/194726 A1   & 42 \\
& WO 2005/029390 A1   & 25 \\
\hline
EP 3 593 317 A1 & WO 2016/025631 A1  & 4\\
& US \, 2015/156369 A1   & 1 \\
\hline
EP 3 382 640 A1 & WO 2015/185944 A1   & 242 \\
& US \, 2005/193205 A1   & 18 \\
& US \, 2016/328398 A1  & 345 \\
\hline
EP 3 438 918 A1 & US \, 2014/029812 A1   & 1 \\
\hline
EP 3 499 493 A1 & US \, 2015/123887 A1   & top score, but... \\
\hline
\end{tabular}
\caption{Reference patents are given in the first column, the search report cited X patents in the second column, the hierarchy position given by BERT in the third column. BERT searched in 1011 patents. Position 1 in column 3 means, that BERT gave the patent in column 2 the highest relevance score of all 1011 patents regarding the first claim of the reference patent in column 1. The details to patent EP~3~499~493~A1 can be found in chapter \ref{discussion}.}
\label{t_results}
\end{table}%

\subsubsection{Discussion}
\label{discussion}

\noindent
The pre-test yielded a F1 score of 0.936. That means, that BERT very well identified the correct descriptions of the claims. To emphasize once again, BERT identified the individual description pieces independently of one another with a very high degree of accuracy, while at the same time recognizing that the description pieces of the other patents are not assigned to the claim. This result gives reason to be confident that BERT can also be used to identify description pieces relevant to a claim in other patents or other documents of the non-patent literature and thus applied successfully to patent novelty searches. Such a novelty search was tested on five patent applications according to table \ref{t_results}. The relevance scoring shows that BERT rates some of the cited X patents as well as highly relevant. The position, given in the third column in table \ref{t_results}, shows the position of the cited X patents in the sample of the 1'011 patents according to the relevance score for the reference patents in column 1. E.g. the X patent US~2015/156369~A1 is on position 1 for the reference patent EP~3~593~317~A1. In the sample of 1'011 patents, this patent was identified by BERT as the most relevant. On the other hand, there are X patents with low rating, like US~2016/328398~A1 on position 345 for reference patent EP~3~382~640~A1. There could be several reasons. In the test, we searched only the first claim. Possibly this patent has been selected as X patent due to at least some of the dependent claims that BERT didn't searched. The X patent US~2015/123887~A1 for reference patent EP~3~499~493~A1 was high rated by BERT as well. But some 10 patents were highly rated as well and indistinguishable from the X patent in terms of the relevance score. This can be explained by the broad claim, which is as follows: "A method of displaying images at a display area of a display device, the method comprising: displaying visible content at a central portion of the display area; and simultaneously displaying invisible content only at one or more edges of the display area." For this reason, the better strategy here may be to include dependent claims for the novelty search as well. 
There are some differences that our search differs from a "real" novelty search. 1: We searched only the first claim. The dependent or further independent claims were not searched. 2: We did not take the priority, application and publication dates into account. The effect could be, that the relevance score of the cited X patents, compared to the other analyzed patents are lower, since patents published after the priority/application date could be of high relevance. 3: The sample of 1'011 patents  were randomly chosen in the same technology field /patent class of the reference patents (with the exception of some X patents). In a real search, BERT's analysis could cover the entire patent class or patents, filtered according to keywords etc. However, it is possible, that a complete consideration of the entire patent class would have worsened the X documents in their position in the scoring. 4: As described above, some of the training and the to-be-searched patents have been truncated due to limits in Excel, whereas the X documents were completely analyzed. It should be examined how these deviations influence the result.
It must also be emphasized that the method proposed here will probably not work if the text pieces, which are of novelty relevance, appear only in a very isolated manner in an analyzed patent. In this case, a relevance scoring by BERT according to the described method could be interpreted as noise. \\

\section{Conclusion and Future Research}
\label{conclusion}

\noindent
In this work, we presented a new method to train BERT for a patent novelty search, by concatenating patent claims to their own descriptions (claim-to-description-BERT). The descriptions of the patents are sliced into description pieces of a certain length, which should be adopted to the length of the trained or searched claim. In our tests, we sliced the description into pieces with a random length in the range of 100 to 200 tokens. We applied the trained BERT in a pre-test on 100 patents, where BERT had to identify the descriptions to a corresponding claim of the same patent. There we got an F1 score of 0.936. We finally applied the trained BERT to a patent novelty search for five patent applications and compared the result to the corresponding search reports. The results showed, that BERT could identify some cited X patents as highly relevant out of a group of 1'011 patents in the same technology field. We have identified some possibilities on how this work can be continued:

\begin{itemize}
\item[1:] Length of the text pieces: We expect the best result when the description pieces are at least as long as the claim or even longer. Finally, the claims include the technical features in a very compact form. A description is particularly relevant if it has all the features of the claim. It is therefore to be expected that a piece of a description of relevance will have at least the same length as the corresponding claim. For training BERT, this could mean dynamically adapting the length of the description to the respective claim and not, as in the test presented here, with a general length or general length interval.
\item[2:] Truncation of the description by Excel. How do these deviations influence the result?
\item[3:] It would be interesting to compare the alternative relevance scoring according to chapter \ref{SRS}.
\item[4:] We trained BERT particularly on B patents (granted patents). The independent claims of B patents tend to have more features than independent claims of the corresponding applications (A patents). The question arises whether BERT should not be better trained on A patents, if the novelty search is applied to A patents. A mixture of A and B patents for training may also be the best strategy. 
\item[5:] We have taken the approach of training BERT technically very close to the features to be searched. However, it may be sufficient here to train the rough technical environment, which would simplify the selection of training patents.
\item[6:] We trained and applied the method only to the first claims. An adoption to dependent claims would be very instructive.
\item[7:]  Extending the described method to search patents to attack the inventive step of a reference patent.\\
\end{itemize}

\section{Acknowledgment}
\noindent
We would like to thank Jochen Spuck, EconSight (Switzerland, Basel) and Carsten Guderian, PatentSight (Germany, Bonn), for their helpful and valuable discussions and support on patent data.

\newpage
\bibliography{Freunek_Bodmer_Claim_to_Description}

\end{document}